\pdfoutput=1

\documentclass[11pt]{article}

\usepackage{acl}
\usepackage{cleveref}

\usepackage{times}
\usepackage{latexsym}
\usepackage{pifont}
\usepackage{bbm}
\usepackage[T1]{fontenc}
\usepackage{multirow}

\usepackage{rotating}

\usepackage{tabularx}
\usepackage[utf8]{inputenc}
\usepackage{booktabs}
\usepackage{microtype}
\usepackage{makecell}

\usepackage{inconsolata}

\title{Generating Attractive and Authentic Copywriting from Customer Reviews}

\author{Yu-Xiang Lin \\
  Academia Sinica \\
  \texttt{yuxianglin@iis.sinica.edu.tw} \\\And
  Wei-Yun Ma\thanks{Corresponding Author} \\
  Academia Sinica \\
  \texttt{ma@iis.sinica.edu.tw} \\}

\begin{document}
\maketitle
\begin{abstract} 

The goal of product copywriting is to capture the interest of potential buyers  by emphasizing the features of products through text descriptions. As e-commerce platforms offer a wide range of services, it's becoming essential to dynamically adjust the styles of these auto-generated descriptions. Typical approaches to copywriting generation often rely solely on specified product attributes, which may result in dull and repetitive content. To tackle this issue, we propose to generate copywriting based on customer reviews, as they provide firsthand practical experiences with products, offering a richer source of information than just product attributes. We have developed a sequence-to-sequence framework, enhanced with reinforcement learning, to produce copywriting that is attractive, authentic, and rich in information. Our framework outperforms all existing baseline and zero-shot large language models, including LLaMA-2-chat-7B and GPT-3.5\footnote{In this work, we use gpt-3.5-turbo-0613. }, in terms of both attractiveness and faithfulness. Furthermore, this work features the use of LLMs for aspect-based summaries collection and argument allure assessment. Experiments demonstrate the effectiveness of using LLMs for marketing domain corpus construction. The code and the dataset is publicly available at:  \url{https://github.com/YuXiangLin1234/Copywriting-Generation}.

\end{abstract}

\section{Introduction}
Copywriting is essential in E-commerce, as it assists online stores in showcasing and marketing their products to prospective buyers considerably. Research has shown that systems capable of automatically generating persuasive copywriting can significantly improve performance metrics on e-commerce platforms \citep{copywriting, guo2022intelligent, controlable-copywriting}, reflecting the value of the automated copywriting creation solutions.

Typical methods of copywriting composition tend to focus only on the predefined attributes of a product \cite{product-description-generation,  copywriting, guo2022intelligent, controlable-copywriting}, leading to content that might be monotonous and unengaging. Instead, testimonials have been proven to induce greater interest among readers \citep{zulkifly2014persuasion},  indicating the effectiveness of copywriting incorporating diverse and authentic customer use cases. Inspired by these viewpoints, our approach emphasizes creating engaging and reliable copywriting from customer reviews. These reviews provide valuable, firsthand insights about the product, thus offering a more detailed and informative perspective than mere product attributes. 

To boost the persuasiveness of produced narration,  
our work endeavors to integrate various dimensions such as factual accuracy, allure in sentence composition, and the richness of information in copywriting. Achieving these objectives simultaneously in a supervised manner presents significant challenges. In contrast, reinforcement learning (RL) algorithms have demonstrated proficiency in handling complex tasks, particularly in the areas of summarization \citep{RLHF-summarization}  and text style transfer \citep{gong-etal-2019-reinforcement}. Our work represents a synthesis of these two tasks, driving us to leverage RL techniques to attain the goal.

We apply RL algorithms with three dedicated but complementary reward models, responsible for attractiveness, faithfulness, and information density,  respectively. For attractiveness, we propose to learn a sentence allure evaluator from pairwise comparisons adjudicated by GPT-3.5\footnote{https://chat.openai.com/}. Our experiments indicate that using the win rate as a metric from these pairwise comparisons to train a simple regression model aligns more closely with ground truth compared to the binary classification model, which is the method typically used in studies focusing on learning human preferences \citep{IBM-convincing, RLHF-2019-openai, RLHF-summarization}. 

Besides, we find that focusing solely on attractiveness in sequence generation might compromise the fidelity, which has been observed in attractive  headline generation studies \citep{Attractive-AAAI2020, honestbait}. To address this, we incorporate a textual entailment model to improve the faithfulness of produced copywriting. Also, to enrich the information of created copywriting, we employ a natural language inference model fine-tuned on the extensive QNLI dataset \citep{glue}, helping to promote the meaningful content and avoid the appearance of gorgeous but meaningless arguments. Our experiments demonstrate that our framework surpasses the baseline in performance, effectively balancing attractiveness, faithfulness, and information richness in the produced content.

In sum, our contributions are mainly threefold:
\begin{itemize}
    \item Our framework introduces an innovative approach to creating compelling, genuine, and information-dense copywriting. This method derives content from off-the-shelf customer reviews, instead of relying on seller-provided product materials. 
    \item We discover that optimizing our reward model based on win rates achieves greater congruence with human judgment compared to binary classification optimization.
    \item We develop a new dataset focused on restaurant review summarization equipped with allure assessment scores with the help of GPT-3.5, reflecting the effectiveness of using LLMs for marketing domain dataset construction. 
\end{itemize}

\section{Related Work}
\subsection{Copywriting Generation}
Previous work generates product copywriting based on product attributes or descriptions \citep{product-description-generation, copywriting, controlable-copywriting}. \citet{product-titles} treat the product title generation task as a named entity recognition problem. \citet{personal-copywriting} build an encoder-decoder framework to generate personalized product descriptions by integrating information about product aspects, user categories, and external knowledge base.

Besides, some work aims to mine the opinions among customer reviews. \citet{Aspect-Opinion-Summarization-1} introduces a CNN-based classifier to select reviews based on pre-defined aspects and group them by sentiment. \citet{aspect-hotel-reviews} extracts sentences in reviews based on similarity with given keywords by Latent Dirichlet Allocation modeling \citep{LDA} but does not focus on the attractiveness. Compared to existing work, we first build an automatic copywriter to create attractive and authentic copywriting, based on customer reviews. 

\subsection{Attractive Arguments Generation}

Many studies have been conducted with the aim of enhancing the appeal of generated sequences. \citet{clickbait} train a CNN-based sensationalism scorer to make Pointer Generator \citep{Pointer-Generator} create sensational headlines. \citet{Attractive-AAAI2020} train a popularity predictor from the click of view and take the popularity score and ROUGE-L score \cite{rouge} to improve the model. A follow-up study claims that the number of clicks may be affected by trending topics, making click rate not a suitable popularity indicator \citep{honestbait}. For example, news about politicians may receive much interest during the election. They alleviate this issue by introducing a writing technique called forward-reference (FR) \cite{forward-reference} into the headline generator. FR aims to motivate the reader's curiosity so that they want to learn more, which is also implicitly employed in earlier research on generating appealing headlines in the form of questions \citep{question-headline-generation}.

Despite their remarkable achievement, the concepts of FR cannot be directly applied to our use case. The writing technique aims to capture the reader's attention, while what we focus on is persuasive copywriting to attract potential buyers. In fact, the most common popularity indicator coming with customer reviews is the star rating. However, a single review may encompass opinions of multiple products or aspects, with the star ratings generally reflecting the overall satisfaction of customers with the shop, making it an inappropriate attractiveness evaluator for individual aspect. 

\subsection{Aspect-based Summarization}
Text summarization has been a focal point of NLP research for a long time. Numerous researchers dig into this field and develop multifarious methods to sum up long documents automatically \citep{summarization-survey-1, summarization-survey-2}. Aspect-based summarization is usually considered when extracting information tailored to different aspects from customer feedback \citep{aspect-based-summarization-source-1}. Summarizing reviews in an abstract fashion may be more appropriate due to the colloquial format and potential spelling errors in customer feedback. \citet{aspect-summarization-structure}  modify the attention mechanism to ensure the model focuses on the information about target keywords. CTRLsum \citep{ctrlsum} builds a framework for automatic keyword extraction during summarization . They find that the pretending aspect of the source document generates summaries related to this aspect. 

Considering that a single review may touch upon numerous aspects, there's a need for aspect-focused copywriting, which is particularly significant in certain scenarios (such as discussing "dishes" in the context of restaurant reviews from Yelp). This requirement for aspect-specific copywriting can be viewed as an unique use case of aspect-based summarization technology. In this work, we adopt the method mentioned in CTRLsum for aspect-based control.

\section{Dataset}
\label{sec:dataset}
Regarding the abstractive summarization benchmark, previous work has predominantly focused on the CNN/Daily Mail dataset \citep{cnn-daily-mail} and the XSum dataset \citep{xsum}. These datasets comprise articles on a broad spectrum of topics and lack relevance to business content, making them unsuitable for our purposes. Additionally, there are lots of existing datasets comprised of real-world customer reviews. Yelp platform provides a large-scale dataset\footnote{https://www.yelp.com/dataset} encompassing customer feedback toward a wide range of business entities. Amazon platform offers multi-language review collections sourced from their e-commerce platform \citep{amazon-dataset}. Nonetheless, these datasets are primarily composed of reviews without corresponding summaries. The absence of a corpus poses a challenge in producing marketing content.

Recently, numerous studies show that summaries generated by LLMs such as GPT-3.5 or 
LLaMA-2 \citep{llama2}, are on par with human-produced summaries \citep{llm-summarization-1, chatgpt-llm-benchmark, chatgpt-multitask}. Also, \citet{chatgpt-aspect-summarization-limitation} shows the remarkable capabilities of GPT-3.5 to conduct text summarization conditioned on given keywords. Built upon these works, we leverage GPT-3.5 to address the lack of corpus.

First, we sample 3622 restaurant reviews from Yelp dataset and then leverage the off-the-shelf keyphrase extraction tool\footnote{https://huggingface.co/ml6team/keyphrase-extraction-distilbert-inspec} to draw out aspects. The tool is built on the methods utilizing contextualized embeddings \citep{keyphrase-2, keyphrase-1}. After that, we ask GPT-3.5 to write summaries based on specified aspects and individual customer reviews. These samples are split into training/validation/testing sets in the ratio of 7:1:2.

\begin{figure}[!tp]
\includegraphics[width=0.45 \textwidth]{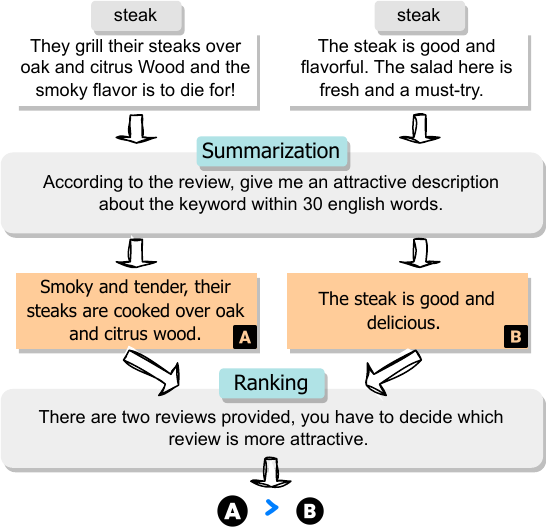}
\caption{The illustration for dataset construction. For simplicity, the requirements of the response format are skipped. We initially replaced "description" with "summary", resulting in a collection of monotonous text.}
\label{fig:dataset_pipeline}
\end{figure}

After generating summaries of reviews, we prompt GPT-3.5 to rank summaries regarding the same aspect. We separate arguments belonging to different splits, ensuring that there was no comparison between two different groups. We find some comparison results violate transitive relation, so a cycle breaking technique \citep{break-cycle} is introduced to clean the dataset. The final version of the ranking dataset consists of 22602/408/1292 samples in training/dev/testing set. 
The pipeline for dataset construction is illustrated in Figure \ref{fig:dataset_pipeline}, with the prompts we used for GPT-3.5.

\section{Methodology}

As previous work indicates the proficiency of RL algorithms in handling complex tasks \citep{RLHF-summarization, gong-etal-2019-reinforcement}, especially the success of Reinforcement Learning from Human Feedback (RLHF) \citep{RLHF-2019-openai, RLHF-summarization, RLHF-diverse, instruct-gpt}, we leverage RL to enhance our model, aiming to align the created copywriting to human preference. 

A typical pipeline of RLHF comprises 3 steps: (1) supervised fine-tuning, (2) reward modeling, and (3) reinforcement learning. We follow these steps as previous work suggests.

\begin{figure*}[!tp]
\centering
\includegraphics[width=\textwidth]{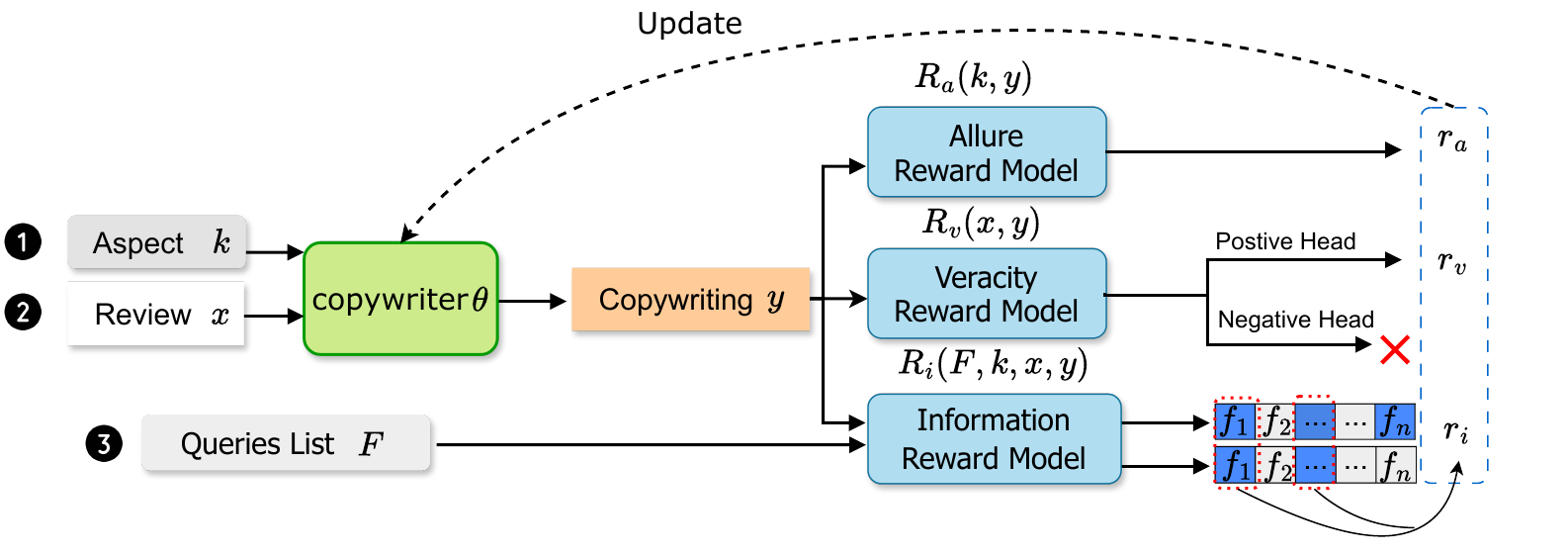}
\caption{The illustration of the proposed framework. Given an aspect $k$, a review $x$, and a list of queries  $F$, we enhance the supervised finetuned model $\theta$ by the Proximal Policy Optimization (PPO) algorithm with three dedicated reward models.}
\label{fig:framework}
\end{figure*}

\subsection{Supervised Fine-tuned Copywriter}

We begin with a pre-trained BART-Large model \cite{bart}. To steer the model toward a proficient copywriter, we fine-tune model $\theta$ by maximizing the likelihood of reference summaries $\hat{y}$ conditioned on source review $x$ and designated aspect $k$. That is, minimizing the cross-entropy loss at each decoding step as
\begin{equation}
    L(\theta) =  -\frac{1}{n} \sum\limits_{i=1}^n \log p(\hat{y_{i}} | \hat{y}_{1:i-1}, x, k, \theta).
    \label{eq:sft}
\end{equation}

\subsection{Reward Modeling}
To further enhance the quality of generated copywriting through reinforcement learning, we leverage the following reward models to conduct argument assessment from multiple perspectives:

\paragraph{Allure Reward} 
First, we build an allure evaluator to gauge the attractiveness of generated copywriting $y$ conditioned on the given aspect $k$. Conceptually, allure assessment determines the attractiveness of arguments based on the writing style, word choice, and persuasiveness.

We learn the allure reward model from binary annotation from GPT-3.5 mentioned in Section  \ref{sec:dataset}.
Related work takes learning pairwise comparison as a binary classification task \citep{IBM-convincing, RLHF-summarization}. Inspired by them, we initially built a Siamese network as our allure reward model and optimized it using cross-entropy classification loss, but found that this baseline did not yield ideal results on the testing set. Hence, we turn to compute the win-rate $w_{i}$ from all $M$ pairwise annotation $\hat{y}_{i}$ engage in, as the alluring score of $\hat{y}_{i}$. The formula is shown as
\begin{equation}
    w_i = \frac{\Sigma_{j=1}^M v_{ij}}{M} ,
    \label{eq:win-rate}\\
\end{equation}
where $v_{ij} = 1$ if $\hat{y_{i}}$ is more attractive than ${\hat{y_{j}}}$ and otherwise $0$ . 

We then learn a regression model to predict the win rate and rank arguments according to their win rate, from highest to lowest. The method significantly improves binary classification accuracy and reduces root mean square error (RMSE). The experimental results are shown in Section \ref{sec:experiment_allure_reward_model}. We adopt this regression model as our allure reward model $R_{a}$, and take the predicted win rate as the allure reward 
    
\begin{equation}
    r_{a} = R_{a}(k, y)
\end{equation}

\paragraph{Veracity Reward} 
Previous work shows there is a high correlation between 
entailment and faithfulness \citep{nli-faithfulness, honestbait}. Inspired by them, we leverage a DeBERTa-v3 model \citep{zeroshot-nli} trained on 33 datasets such as WANLI \citep{wanli} and MultiNLI \citep{mnli}, and thus owning rich knowledge and remarkable zero-shot capabilities, to serve as our veracity reward model. In this work, we adopt the model provided by the authors\footnote{https://huggingface.co/MoritzLaurer/deberta-v3-large-zeroshot-v1.1-all-33}. Since the model outputs two values representing "entailment" and "not entailment," we choose the output logits of the "entailment" head as our veracity reward

\begin{equation}
    r_{v} = R_{v}(x, y).
\end{equation}

The statement refers to estimating whether review $x$ logically entails output $y$, with an objective to compose copywriting that accurately reflects the source review without any misrepresentation.

\paragraph{Information Reward}
Intuitively, the copywriter optimized with the above two rewards might generate gorgeous but pointless template-based arguments such as "The food is delicious and perfect." or "The food is a mouthwatering delight that will leave you craving for more." To tackle this issue, we encourage the copywriter to provide more information, like commendable characteristics or the difference between similar products on the market.  We leverage a model trained on a large-scale dataset named QNLI \citep{glue} to attain the purpose. QNLI dataset consists of question-paragraph pairs, and labels indicate whether the answer to the question can
be found in the paragraph. With some common facets of restaurant reviews, we query the QNLI model to check whether information about these facets is in the generated copywriting. Additionally, to ensure the fidelity to the source review, we also check what the fact review contains. The information reward is formulated as
\begin{equation}
\label{information_reward}
    r_i = \frac{|\Sigma_{f \in F} R_{i}(f, k, x) \land R_{i}(f, k, y)|}{|\Sigma_{f \in F} R_{i}(f, k, y)|} ,
\end{equation}

where $|\cdot|$ means the counting function, $F$ refers to the set of queries about different facets, such as “How does $k$ look?” or ”How does $k$ compare to similar ones on the 
market?”. Our queries are provided in Appendix \ref{sec:queries}. Conceptually, if the information is included in source documents, it should be in generated summaries. In this case, there might not be a difference between an attractive sentence and an unappealing one. But with the help of the other two RMs, we are able to create a proficient copywriter producing persuasive copywriting.

\subsection{Reinforcement Learning}

With the supervised fine-tuned model $\theta$, the allure RM $R_{a}$, the veracity RM $R_{v}$, and the information RM $R_{i}$, we further enhance $\theta$ with the Proximal Policy Optimization (PPO) algorithm \citep{ppo}, as illustrated in Figure \ref{fig:framework}. PPO shows the remarkable capability of guiding the policy toward a desired direction with high sampling efficiency. We treat $\theta$ as policy and optimize $\theta$ as 
\begin{equation}
     \mathbbm{R}(x, y, k, F) = \alpha r_{a} + \beta r_{v} + \gamma r_{i} - \log \frac{\theta(y|x, k)}{\theta_{old}(y|x, k)},
    \label{eq:RL}
\end{equation}

where the last term refers to a penalty, $\theta_{old}$ is the original version of $\theta$ before RL optimization. The KL-divergence-based penalty is meant to prevent $\theta$ moving too far from $\theta_{old}$ and thus stabilize the training. $\alpha$, $\beta$ and $\gamma$ are weight of different rewards. For simplicity, we set $\alpha$, $\beta$ and $\gamma$ to be 1 in all following experiments.

\section{Experiments}
All experiments are conducted on an Nvidia Tesla-V100 GPU. We select the model that performs the best on a validation set. The implementation details are listed in Appendix~\ref{sec:implementation_details}.

\subsection{Allure Reward Model}
\label{sec:experiment_allure_reward_model}
We initially built a Siamese network as our allure reward model and optimized it using cross entropy classification loss as suggested in related work \citep{IBM-convincing, RLHF-summarization}. However, we find that this baseline does not yield ideal results on the testing set. We turn to learn a regression model with the win rate for each argument and get a better result. The experiments are conducted on a BART-base model \citep{bart} and DeBERTa-v3-base model \citep{deberta-v3}, with searching learning rate in $[1e-5, 3e-5, 5e-5, 7e-5, 9e-5]$. we report the best results about root mean square error (RMSE) and binary classification accuracy on testing set in Table \ref{tab:allure_reward}.

It's reasonable that the regression model exhibits a lower RMSE metric than the Siamese network since RMSE is computed based on win rate. However, the gap between these two types of models is too large despite the existence of a sigmoid layer. Surprisingly, the accuracy of the DeBERTa-v3 regression model is much higher than that of the Siamese network. According to the experiments, we select the DeBERTa-v3 regression model, which exhibits the best performance, as our allure reward model in the following experiments.

\begin{table}[h]
\scalebox{1}{
\begin{tabular}{c|lcc}
\toprule[1pt] 
\midrule
    & Model      & Acc. $\uparrow$ & RMSE $\downarrow$ \\ \midrule
\multirow{2}{*}{\rotatebox[origin=c]{90}{\tiny\textbf{BART}}} 
    & Siamese Network       & 66.25 \%     & 0.52 \\
    & Regression Network & 66.25 \%  & \textbf{0.37}     \\ 
    \midrule \cmidrule{1-1}
\multirow{2}{*}{\rotatebox[origin=c]{90}{\tiny\textbf{DeBERTa}}}     
    & Siamese Network     & 65.09 \% & 0.54   \\
    & Regression Network  & \textbf{69.50} \% & \textbf{0.37}  \\
\midrule 
\bottomrule[1pt]
\end{tabular}}
\caption{Experiment about allure reward model}
\label{tab:allure_reward}
\end{table}

\subsection{Baselines for Copywriting Generation}
We compare our proposed framework to the following baselines. We choose works built on a BART-large model to alleviate the effect of different backbones. Summaries in our dataset are both attractive and faithful to source review inherently since we prompt GPT-3.5 to generate such samples. Hence, baseline fine-tuned on them should highlight their strong abilities to generate attractive and authentic copywriting. We first compare our model to the supervised fine-tuning (\textsf{SFT}) model for checking the effect of reinforcement learning. Besides, an aspect-based summarization model called \textsf{CTRLsum}  \citep{ctrlsum} shows good controllability to paraphrase the sentence based on given aspects. Some work utilizes ROUGE score to enhance the quality of the abstract summarization and achieve notable success \citep{rouge-rl-summarization-1, rouge-rl-summarization-2}, making us put the model optimized by ROUGE reward into comparison. We compute ROUGE-L score as a reward for training this baseline, as suggested in their paper. 

We also incorporate LLMs in comparison since LLMs demonstrate remarkable zero-shot capabilities on multiple tasks \citep{chatgpt-llm-benchmark, llm-summarization-1}. Given a keyword $k$ and source review $x$, we prompt \textsf{LLaMA-2-chat-7B} \citep{llama2} to generate attractive and faithful summaries, serving as one of the baselines. The prompt employed here is the same as the summarization part in Figure \ref{fig:dataset_pipeline} and adjusted to the suggested format of LLaMA-2-Chat. The last baseline is summaries generated by \textsf{GPT-3.5}. Note that they are ground truth for supervised fine-tuning.

\subsection{Human evaluation}
\label{human-evaluation-setting}
Since we use the large language model for dataset construction, it's more reliable for conducting human evaluation. In this experiment, we randomly sample 100 customer reviews and the corresponding copywriting from the testing set for comparison. Given a customer review $x$, an aspect $k$, and two copywriting $y_1$, $y_2$, The instructions for evaluators are (1) Which one is more attractive and makes you want to purchase/enjoy $k$? (2) Which one is more faithful to the original review $x$? (3) Which one is more fluent? The evaluators are asked to select \textit{first copywriting}, \textit{second copywriting}, or \textit{tie} for each pair and each question. We give \$0.05 for each pair of data because of the simplicity of annotation. We use evaluators in the United States and the United Kingdom, with a greater than 95\% approval rate.

The evaluation protocol follows previous work \citep{Human-evaluation-setting,honestbait} to evaluate the generated copywriting's attractiveness, faithfulness, and fluency by pairwise comparison. Results are reported in Table \ref{tab:huamn_evaluation_all}. For example, the copywriting generated by our framework is 55.33\%/34.33\%/10.33\%  better than/worse than/equal to the output of the \textsf{SFT} model on faithfulness. The value of faithfulness in the table thus becomes 34.33\% - 55.33\% = -21\%. Results show that our framework outperforms all baselines, including zero-shot LLMs and fine-tuned BART model, in terms of attractiveness and faithfulness, with comparable fluency. This implies the positive impact of RL finetuning. The results also reflects the controllability of our framework because the judgment criteria for attractiveness are based on the given aspect.

\begin{table}[h]
\scalebox{0.82}{
\begin{tabular}{c|lrrr}
\toprule[1pt] 
\midrule
    & Model      & ATRC $\uparrow$ & FAITH $\uparrow$ & FLCY $\uparrow$ \\ \midrule
\multirow{2}{*}{\rotatebox[origin=c]{90}{\textbf{LLM}}} 
    & GPT-3.5        & -10.0\%     & -3.3\%     & 3.0\%  \\
    & LLaMA-2-7B-Chat & -25.0\%  & -19.3\%  & 3.3\%    \\ 
    \midrule \cmidrule{1-1}
\multirow{3}{*}{\rotatebox[origin=c]{90}{\textbf{BART}}}     
    & SFT           & -15.0\% & -21.0\%  & -13.3\% \\
    & CTRLsum          & -17.0\% & -5.3\%  & -2.7\%  \\
    
    & ROUGE             & -13.7\% & -4.7\% & 3.0\%  \\
    & Ours               & - & -  & - \\
\midrule 
\bottomrule[1pt]
\end{tabular}}
\caption{Human evaluation results of pairwise comparison. We compare copywriting generated by our framework and baseline models in terms of attractiveness (ATRC), faithfulness (FAITH), and fluency (FLCY).}
\label{tab:huamn_evaluation_all}
\end{table}

\subsection{Automatical Metrics}
\paragraph{ROUGE Score} ROUGE score \citep{rouge} is a conventional metric to measure the quality of text summarization. It counts word-level overlap to estimate the similarity between the generated summary and the reference one. ROUGE score has been shown to be unconvincing in abstractive summarization task \citep{rouge-limits} . Our framework aims to build a copywriter capable of generating attractive copywriting. Using the ROUGE score to measure the performance is inappropriate, but we still provide ROUGE-1, ROUGE-2, ROUGE-L in Table \ref{tab:rouge} for reference. Most baseline models exhibit comparable ROUGE scores,  except for \textsf{LLaMA-2-7B-Chat}, which may be attributed to the zero-shot generation. \textsf{SFT} achieves the highest ROUGE score, which is reasonable since it closely imitates the writing style of reference summaries, although it yields the worst results comprehensively. Interestingly, the model optimized by the ROUGE reward exhibits a similar ROUGE score with other baselines despite the increasing ROUGE reward during the training process. We think the reason is the broad knowledge \textsf{GPT-3.5} owns, and the dataset is constructed in a zero-shot manner, resulting in a different distribution between the training and testing sets.

\begin{table}[h]
\begin{tabular}{c|lccc}
\toprule[1pt] \midrule
& Model             & $R_1$ & $R_2$ & $R_L$ \\ \midrule
\multirow{2}{*}{\rotatebox[origin=c]{90}{\textbf{LLM}}} 
    & GPT-3.5 (gold)          & -     & -     & -     \\
    & LLaMA-2-7B-Chat & 18.50  & 3.55  & 15.00    \\ 
    \midrule \cmidrule{1-1}
\multirow{4}{*}{\rotatebox[origin=c]{90}{\textbf{BART}}}     
    & SFT           & \textbf{28.25} & \textbf{8.13}  & \textbf{23.94} \\
    & CTRLsum           & 27.22 & 7.52  & 23.30  \\
    
    & ROUGE             & 27.26 & 7.63  & 23.40  \\
    & Ours               & 27.28 & 7.34  & 23.18 \\

    \midrule \bottomrule[1pt]
\end{tabular}
\caption{ROUGE score computed with reference summary generated by GPT-3.5. $R_1$, $R_2$, $R_L$ are the ROUGE-1, ROUGE-2, ROUGE-L score respectively.}
\label{tab:rouge}
\end{table}

\paragraph{Perplexity}
Perplexity (PPL) is commonly used to evaluate text fluency \citep{perplexity-fluency-1, perplexity-fluency-2}. The metric refers to the likelihood of pre-trained language models (PLMs) to produce the sequence. Lower PPL means better fluency of the sentences. We compute PPL in Table \ref{tab:ppl} with two publicly available PLMs, GPT-2 \citep{gpt-2} and LLaMA-2-7B \citep{llama2}. Results show that our framework achieves the lowest PPL, the highest fluency. 

\begin{table}[h]
\scalebox{0.9}{
\begin{tabular}{c|lcc}
\toprule[1pt] 
\midrule
    & Model      & GPT-2 & LLaMA-2-7B  \\
    \midrule
\multirow{2}{*}{\rotatebox[origin=c]{90}{\textbf{LLM}}} 
    & GPT-3.5      & 67.10     & 32.59     \\
    & LLaMA-2-7B-Chat & 71.88  & 46.36    \\ 
    \midrule \cmidrule{1-1}
\multirow{4}{*}{\rotatebox[origin=c]{90}{\textbf{BART}}}     
    & SFT               & 49.75 & 26.86   \\
    & CTRLsum           & 48.93 & 26.89   \\
    & ROUGE             & 52.05 & 26.10   \\
    & Ours              & \textbf{48.63} & \textbf{25.23}  \\
    
\midrule 
\bottomrule[1pt]
\end{tabular}
}
\caption{Perplexity (PPL) computed by GPT-2 and LLaMA-2-7B. Lower PPL means better fluency. }
\label{tab:ppl}
\end{table}

\subsection{Ablation Study}
We conduct the following ablation analyses to probe the impact of components in our framework.

\paragraph{Human Evaluation}
We repeat the experiments in Section \ref{human-evaluation-setting}. Since the enhancement of RL over the SFT model has been shown in previous experiments, we learn three models removing allure RM $R_{a}$, veracity RM $R_{v}$, and information RM $R_{i}$ in this section. Results are demonstrated in Table \ref{tab:ablation-human}. Intuitively, models optimized without $R_{a}$ and  $R_{v}$ yield the degradation of attractiveness and faithfulness. The information reward model seems to be helpful for faithfulness. We claim that the purpose of $R_{i}$ is to increase the information density of generated copywriting that the source document mentioned. Hence, with the help of $R_{v}$ and $R_{i}$, copywriters tend to offer customers more accurate information about different facets. In this case, copywriting with many "facts" will be more faithful to that with only empty words. 
  
\begin{table}[h]
\centering
\begin{tabular}{lrrr}
\toprule[1pt] 
\midrule
          & ATRC $\uparrow$ & FAITH $\uparrow$ & FLCY $\uparrow$ \\
    \midrule
    Ours               & - & -  & - \\
    \quad  W/o $R_{a}$   & -14.0\% & -0.0\%  & -5.7\% \\
    \quad  W/o $R_{v}$      & -8.3\% & -14.0\%  & -6.7\% \\
    \quad  W/o $R_{i}$  &  2.0\% & -8.7\%  & -2.3\% \\
\midrule 
\bottomrule[1pt]
\end{tabular}
\caption{Ablation study with human evaluation in terms of attractiveness (ATRC), faithfulness (FAITH), fluency (FLCY).}
\label{tab:ablation-human}
\end{table}

\paragraph{Information Score}
In Table \ref{tab:inforamtion_score}, we compute the information score of generated summaries in the testing set as Equation \ref{information_reward} by QNLI model. The experiment is to further confirm the impact of the information reward model $R_{i}$. Additionally, whether the information score is proportional to the sentence length is questionable. We write down the mean and standard deviation of unigram counts for each baseline. All baseline models generated copywriting within 30 words, as required in Section \ref{sec:dataset} for generated SFT training data by GPT-3.5. Results show that our framework achieves the highest information reward with around 3 words longer. Instead, the model optimized without information reward produces longer sentences on average but with the lowest information score, verifying the significance of $R_{i}$. 

\begin{table}[h]
\scalebox{0.9}{
\begin{tabular}{c|l|rr|c}
\toprule[1pt] 
\midrule
    & Model      & Avg. & Std.   & INFO $\uparrow$ \\ \midrule
\multirow{2}{*}{\rotatebox[origin=c]{90}{\textbf{LLM}}} 
    & GPT-3.5     & 19.09     & 6.34    & 56.45  \\
    & LLaMA-2-7B-Chat & 17.03  & 12.04  & 55.42   \\ 
    \midrule \cmidrule{1-1}
\multirow{5}{*}{\rotatebox[origin=c]{90}{\textbf{BART}}} 
    & SFT               & 19.14 & 5.81 & 57.57  \\    
    & CTRLsum           & 15.71 & 4.81 & 56.79  \\
    
    & ROUGE             & 20.55 & 7.38 & 56.48  \\
    & Ours              & 22.57    & 8.00   & \textbf{61.06}\\
    & \quad W/o $R_{i}$ & 23.28    & 8.55   & 56.48 \\
\midrule 
\bottomrule[1pt]
\end{tabular}
}
\caption{Length of generated sentences and information score. Avg. and Std. refer to the average counts of uni-grams and the standard deviation of sentence length, respectively. INFO refers to the information score computed by QNLI model.}
\label{tab:inforamtion_score}
\end{table}

\subsection{Qualitative Results}
As an example, we randomly select a sample with two aspects in the testing set. Table \ref{tab:generated_example_part} shows the source review and corresponding copywriting. Due to the page limit, we only list the output of the two strongest baselines in the above experiments, \textsf{GPT-3.5} and \textsf{CTRLsum}. Copywriting generated by other baselines is listed in Table \ref{tab:generated_example_full}. For both aspects, "Steak" and "Tampa," we can see our copywriter provides more information mentioned in the source review, such as "Sides were veggies and sweet potatoes" and "delicious veggies and sweet potatoes." And for the aspect "Tampa", \textsf{GPT-3.5} and \textsf{CTRLsum} say that there is a diverse option for meals, but our framework lists what the dining options include. For attractiveness, we find that \textsf{GPT-3.5} tends to produce gorgeous but advanced words, making it hard to attract turker at first glance. This might be the reason why our framework performs better than \textsf{GPT-3.5} regarding attractiveness and faithfulness.  

\section{Comparisons with Human-written Copywriting}

We have taken the following steps to bolster the experiments with some human-written copywriting to address concerns regarding using LLMs for training data collection and the word length constraints applied to LLM baselines.

We employ Amazon Mechanical Turk workers to create copywriting samples. Specifically, we gather 100 human-written copywriting pieces from testing samples used in Section \ref{human-evaluation-setting}. To ensure the quality of these human-generated samples, we meticulously filter out 20 significantly inappropriate or off-topic responses. The instructions given to the turkers are as follows: \textit{Please write an attractive but faithful summary based on the given keyword and customer review. The summary will be used as product copywriting. You should paraphrase the sentence to make it more attractive and attract potential buyers to buy/enjoy this keyword. But it should be faithful to the original review.}
\subsection{Human Evaluation}
After incorporating the human-written summaries, we conduct a human evaluation with these samples regarding three key aspects, the same as conditions outlined in Section \ref{human-evaluation-setting}. The comparative results are shown in Table \ref{tab:huamn_evaluation_human}, showcasing that copywriting generated by our framework is superior quality compared to human writers.

\begin{table}[h]
\scalebox{0.95}{
\begin{tabular}{lrrr}
\toprule[1pt] 
\midrule
      & ATRC $\uparrow$ & FAITH $\uparrow$ & FLCY $\uparrow$ 
    \\ 
    \midrule
    Human-written     & -2.09\%     & -0.81\%     & -6.25\%  
    \\
    Ours   & - & -  & - \\
\midrule 
\bottomrule[1pt]
\end{tabular}}
\caption{Human evaluation results of pairwise comparison. We compare copywriting generated by our framework and human-written one in terms of attractiveness (ATRC), faithfulness (FAITH), and fluency (FLCY).}
\label{tab:huamn_evaluation_human}
\end{table}

\subsection{Will Language Models Generate Sentences Humans Will Produce?}
To further alleviate the concern about the ability of LLMs to produce copywriting as persuasive and attractive as human authors, we employ Fast-detectgpt \citep{fastdetectgpt}, a state-of-the-art toolkit for machine-generated text detection, to verify the human-like quality of the copywriting generated by our framework. Fast-detectgpt is proficient in detecting machine-generated text in a zero-shot manner, making it an ideal tool for this assessment.

Adopting Fast-detectgpt, we analyze the copywriting produced by our framework to determine the likelihood of it being identified as machine-generated. The results presented in Table \ref{tab:detectgpt} indicate low probabilities of the text being flagged as machine-generated or 'Fake.' This implies that the text generated by our model closely resembles human-written content, both in style and substance.

\begin{table}[h]
\centering
\scalebox{1}{
\begin{tabular}{lc}
\toprule[1pt] 
\midrule
          & probability $\downarrow$
    \\ 
    \midrule
    Human-written     & 	\textbf{33.22}\%  
    \\
    Ours   & 37.44\%
    \\
\midrule 
\bottomrule[1pt]
\end{tabular}}
\caption{probabilities of the text are machine-generated}
\label{tab:detectgpt}
\end{table}

\begin{table*}[tp]
\centering
\scalebox{0.9}
{
\begin{tabular}{l|>{\arraybackslash}p{7cm}|>{\arraybackslash}p{7cm}}
\toprule[1pt] 
    \midrule
     \multicolumn{1}{c|}{Aspect} & \multicolumn{1}{c|}{Steak} & \multicolumn{1}{c}{Tampa} \\
    \midrule
     Source Review      & \multicolumn{2}{>{\arraybackslash}p{14cm}}{We got diverted to Tampa and decided to make the drive to Orlando, but first we needed somewhere to eat. Wanting to avoid chain restaurants, yelp steered us toward this little local gem! They grill their steaks over oak, and citrus Wood  and the smoky flavor is to die for! We are big fans of our smoker back home  and cook our steaks this way as well. It's rare to find steak cooked over woodsmoke  in most places, and this was worth the side trip!  The sides were veggies and  sweet potatoes. The wines on offer were so reasonably priced - most under \$20 a  bottle, but a decent selection. They just don't mark them up 5x like most places!  Bookmarking this place for a return trip next time I am in Tampa for sure!}   \\ 

    \midrule[1pt]
    \midrule

     GPT-3.5    & Juicy, tender, and perfectly grilled, this oak and citrus wood-smoked steak delivers a mouthwatering  smoky flavor. Cooked to perfection, it's a rare  gem worth savoring, a true delight for steak enthusiasts. &  An enchanting city in Florida known for its diverse dining options, Tampa captivates visitors with its local gems and the tantalizing aroma of woodsmoked steaks
     \\ 

    \midrule
    
     CTRLsum           & Mouthwatering, tender, and smoky steaks grilled over oak and citrus Wood,  bursting with a delectable flavor  that will leave you craving for more. &  Vibrant city with diverse culinary options,  including a local gem serving smoky steaks  and delicious sides.
     \\

    \midrule
    
     Ours              & Juicy, smoky, and grilled over oak and citrus wood, our steaks at this local gem in Tampa are truly exceptional.  A rare delight cooked over woodsmoke,  it's worth the  side trip! The sides were veggies and sweet potatoes.  &  Vibrant Florida city with smoky steaks, affordable wines,  delicious veggies  and sweet potatoes.  A must-visit!    \\
\midrule 
\bottomrule[1pt]
\end{tabular}
}
\caption{The generated examples of different models. Due to the page limit, we provide the copywriting generated by each baseline model in Appendix \ref{app:generated_example_full}.}
\label{tab:generated_example_part}
\end{table*}

\section{Conclusion}

In this paper, we present an innovative framework with an elegant but effective reward mechanism, designed to automatically generate attractive and faithful copywriting. Extensive experiments are conducted to verify the superior capabilities of the proposed method, compared to existing baselines including zero-shot Large Language Models (LLMs), fine-tuned BART models, and even human writers. Moreover, our findings reveal that optimizing the reward model using win-rate leads to a higher alignment with human judgment than the traditional binary classification optimization method. This approach enhances the model's ability to reflect human preferences more accurately.

In addition to these developments, we have created an unique dataset comprising summaries and corresponding ranking data, which we believe will be valuable for future research in this field. Our dataset is built with the help of GPT-3.5 and includes curated content specifically tailored to advancing the capabilities of automatic copywriting systems, demonstrating the effectiveness of using LLMs for marketing domain dataset construction.  Through this work, we aspire to expand the potential and effectiveness of E-commerce platforms, offering new avenues for engaging and authentic content generation.

\section*{Scientific Artifacts}
All experiments are conducted upon PyTorch \citep{PyTorch}, and all pretrained models other than GPT-3.5 are obtained from HuggingFace \citep{Huggingface}. We also use NLTK \citep{NLTK} package for word tokenization and sentence tokenization. We adopt these artifacts for their intended use.

\section*{Limitation and Potential Risks}
Our framework is not applicable when customer reviews are lacking, especially when the product is unavailable on the market. The proposed approach underscores our key point: generating copywriting from customer reviews is a novel and valuable direction. Just as pre-release copywriting sourced seller-defined attributes has merits, post-release revisions based on actual customer experiences can provide a wealth of information beyond mere product attributes. Both pre-release generation and post-release refinement of copywriting are essential, complementing each other in effectively marketing a product. 

Also, despite our framework exhibiting the best attractiveness and faithfulness, there is still room for improvement. For attractiveness, the binary classification accuracy and the agreement with the ground truth of our allure model are around 70 percent. For faithfulness, our framework produces some fake information once in a while. We believe these issues can be alleviated by developing a more accurate reward evaluator. Building upon the rapid development of LLMs,  \citet{gpt3-as-reward} prompts LLMs to compute reward, which might be the direction of our future work.

\section*{Acknowledgements}
We are grateful for the insightful and valuable comments from anonymous reviews. Thanks also to Yin-Hsiang Liao for the preliminary discussions. This work is supported by the National Science and Technology Council of Taiwan under grant numbers NSTC112-2221-E-001-025. Thanks to the National Center for High-performance Computing (NCHC) for providing computational and storage resources.

\bibliography{custom}

\begin{thebibliography}{60}
\expandafter\ifx\csname natexlab\endcsname\relax\def\natexlab#1{#1}\fi

\bibitem[{Akhtar et~al.(2017)Akhtar, Zubair, Kumar, and Ahmad}]{aspect-hotel-reviews}
Nadeem Akhtar, Nashez Zubair, Abhishek Kumar, and Tameem Ahmad. 2017.
\newblock \href {https://doi.org/https://doi.org/10.1016/j.procs.2017.09.115} {Aspect based sentiment oriented summarization of hotel reviews}.
\newblock \emph{Procedia Computer Science}, 115:563--571.
\newblock 7th International Conference on Advances in Computing \& Communications, ICACC-2017, 22-24 August 2017, Cochin, India.

\bibitem[{Allahyari et~al.(2017)Allahyari, Pouriyeh, Assefi, Safaei, Trippe, Gutierrez, and Kochut}]{summarization-survey-2}
Mehdi Allahyari, Seyedamin Pouriyeh, Mehdi Assefi, Saeid Safaei, Elizabeth~D. Trippe, Juan~B. Gutierrez, and Krys Kochut. 2017.
\newblock \href {https://doi.org/10.14569/IJACSA.2017.081052} {Text summarization techniques: A brief survey}.
\newblock \emph{International Journal of Advanced Computer Science and Applications}, 8(10).

\bibitem[{Bakker et~al.(2022)Bakker, Chadwick, Sheahan, Tessler, Campbell-Gillingham, Balaguer, McAleese, Glaese, Aslanides, Botvinick, and Summerfield}]{RLHF-diverse}
Michiel Bakker, Martin Chadwick, Hannah Sheahan, Michael Tessler, Lucy Campbell-Gillingham, Jan Balaguer, Nat McAleese, Amelia Glaese, John Aslanides, Matt Botvinick, and Christopher Summerfield. 2022.
\newblock \href {https://proceedings.neurips.cc/paper_files/paper/2022/file/f978c8f3b5f399cae464e85f72e28503-Paper-Conference.pdf} {Fine-tuning language models to find agreement among humans with diverse preferences}.
\newblock In \emph{Advances in Neural Information Processing Systems}, volume~35, pages 38176--38189. Curran Associates, Inc.

\bibitem[{Bang et~al.(2023)Bang, Cahyawijaya, Lee, Dai, Su, Wilie, Lovenia, Ji, Yu, Chung, Do, Xu, and Fung}]{chatgpt-multitask}
Yejin Bang, Samuel Cahyawijaya, Nayeon Lee, Wenliang Dai, Dan Su, Bryan Wilie, Holy Lovenia, Ziwei Ji, Tiezheng Yu, Willy Chung, Quyet~V. Do, Yan Xu, and Pascale Fung. 2023.
\newblock \href {http://arxiv.org/abs/2302.04023} {A multitask, multilingual, multimodal evaluation of chatgpt on reasoning, hallucination, and interactivity}.

\bibitem[{Bao et~al.(2023)Bao, Zhao, Teng, Yang, and Zhang}]{fastdetectgpt}
Guangsheng Bao, Yanbin Zhao, Zhiyang Teng, Linyi Yang, and Yue Zhang. 2023.
\newblock Fast-detectgpt: Efficient zero-shot detection of machine-generated text via conditional probability curvature.
\newblock \emph{arXiv preprint arXiv:2310.05130}.

\bibitem[{Bird et~al.(2009)Bird, Klein, and Loper}]{NLTK}
Steven Bird, Ewan Klein, and Edward Loper. 2009.
\newblock \emph{Natural language processing with Python: analyzing text with the natural language toolkit}.
\newblock " O'Reilly Media, Inc.".

\bibitem[{Blei et~al.(2003)Blei, Ng, and Jordan}]{LDA}
David~M Blei, Andrew~Y Ng, and Michael~I Jordan. 2003.
\newblock Latent dirichlet allocation.
\newblock \emph{Journal of machine Learning research}, 3(Jan):993--1022.

\bibitem[{Blom and Hansen(2015)}]{forward-reference}
Jonas~Nygaard Blom and Kenneth~Reinecke Hansen. 2015.
\newblock \href {https://doi.org/https://doi.org/10.1016/j.pragma.2014.11.010} {Click bait: Forward-reference as lure in online news headlines}.
\newblock \emph{Journal of Pragmatics}, 76:87--100.

\bibitem[{Chen et~al.(2023)Chen, Wu, and Ku}]{honestbait}
Chih~Yao Chen, Dennis Wu, and Lun-Wei Ku. 2023.
\newblock \href {https://doi.org/10.18653/v1/2023.findings-acl.296} {{H}onest{B}ait: Forward references for attractive but faithful headline generation}.
\newblock In \emph{Findings of the Association for Computational Linguistics: ACL 2023}, pages 4810--4824, Toronto, Canada. Association for Computational Linguistics.

\bibitem[{Chen et~al.(2019)Chen, Lin, Zhang, Yang, Zhou, and Tang}]{personal-copywriting}
Qibin Chen, Junyang Lin, Yichang Zhang, Hongxia Yang, Jingren Zhou, and Jie Tang. 2019.
\newblock \href {https://doi.org/10.1145/3292500.3330725} {Towards knowledge-based personalized product description generation in e-commerce}.
\newblock In \emph{Proceedings of the 25th ACM SIGKDD International Conference on Knowledge Discovery \& Data Mining}, KDD '19, page 3040–3050, New York, NY, USA. Association for Computing Machinery.

\bibitem[{Dathathri et~al.(2020)Dathathri, Madotto, Lan, Hung, Frank, Molino, Yosinski, and Liu}]{perplexity-fluency-2}
Sumanth Dathathri, Andrea Madotto, Janice Lan, Jane Hung, Eric Frank, Piero Molino, Jason Yosinski, and Rosanne Liu. 2020.
\newblock \href {https://openreview.net/forum?id=H1edEyBKDS} {Plug and play language models: A simple approach to controlled text generation}.
\newblock In \emph{International Conference on Learning Representations}.

\bibitem[{Falke et~al.(2019)Falke, Ribeiro, Utama, Dagan, and Gurevych}]{nli-faithfulness}
Tobias Falke, Leonardo F.~R. Ribeiro, Prasetya~Ajie Utama, Ido Dagan, and Iryna Gurevych. 2019.
\newblock \href {https://doi.org/10.18653/v1/P19-1213} {Ranking generated summaries by correctness: An interesting but challenging application for natural language inference}.
\newblock In \emph{Proceedings of the 57th Annual Meeting of the Association for Computational Linguistics}, pages 2214--2220, Florence, Italy. Association for Computational Linguistics.

\bibitem[{Frermann and Klementiev(2019)}]{aspect-summarization-structure}
Lea Frermann and Alexandre Klementiev. 2019.
\newblock \href {https://doi.org/10.18653/v1/P19-1630} {Inducing document structure for aspect-based summarization}.
\newblock In \emph{Proceedings of the 57th Annual Meeting of the Association for Computational Linguistics}, pages 6263--6273, Florence, Italy. Association for Computational Linguistics.

\bibitem[{Gleize et~al.(2019)Gleize, Shnarch, Choshen, Dankin, Moshkowich, Aharonov, and Slonim}]{IBM-convincing}
Martin Gleize, Eyal Shnarch, Leshem Choshen, Lena Dankin, Guy Moshkowich, Ranit Aharonov, and Noam Slonim. 2019.
\newblock \href {https://doi.org/10.18653/v1/P19-1093} {Are you convinced? choosing the more convincing evidence with a {S}iamese network}.
\newblock In \emph{Proceedings of the 57th Annual Meeting of the Association for Computational Linguistics}, pages 967--976, Florence, Italy. Association for Computational Linguistics.

\bibitem[{Gong et~al.(2019)Gong, Bhat, Wu, Xiong, and Hwu}]{gong-etal-2019-reinforcement}
Hongyu Gong, Suma Bhat, Lingfei Wu, JinJun Xiong, and Wen-mei Hwu. 2019.
\newblock \href {https://doi.org/10.18653/v1/N19-1320} {Reinforcement learning based text style transfer without parallel training corpus}.
\newblock In \emph{Proceedings of the 2019 Conference of the North {A}merican Chapter of the Association for Computational Linguistics: Human Language Technologies, Volume 1 (Long and Short Papers)}, pages 3168--3180, Minneapolis, Minnesota. Association for Computational Linguistics.

\bibitem[{Goyal et~al.(2023)Goyal, Li, and Durrett}]{llm-summarization-1}
Tanya Goyal, Junyi~Jessy Li, and Greg Durrett. 2023.
\newblock \href {http://arxiv.org/abs/2209.12356} {News summarization and evaluation in the era of gpt-3}.

\bibitem[{Guo et~al.(2022{\natexlab{a}})Guo, Wang, Zhao, Diao, Chen, Ding, He, Lu, Xiao, Long et~al.}]{guo2022intelligent}
Xiaojie Guo, Shugen Wang, Hanqing Zhao, Shiliang Diao, Jiajia Chen, Zhuoye Ding, Zhen He, Jianchao Lu, Yun Xiao, Bo~Long, et~al. 2022{\natexlab{a}}.
\newblock Intelligent online selling point extraction for e-commerce recommendation.
\newblock In \emph{Proceedings of the AAAI Conference on Artificial Intelligence}, volume~36, pages 12360--12368.

\bibitem[{Guo et~al.(2022{\natexlab{b}})Guo, Zeng, Jiang, Xiao, Long, and Wu}]{controlable-copywriting}
Xiaojie Guo, Qingkai Zeng, Meng Jiang, Yun Xiao, Bo~Long, and Lingfei Wu. 2022{\natexlab{b}}.
\newblock Automatic controllable product copywriting for e-commerce.
\newblock In \emph{Proceedings of the 28th ACM SIGKDD Conference on Knowledge Discovery and Data Mining}, pages 2946--2956.

\bibitem[{He et~al.(2022)He, Kryscinski, McCann, Rajani, and Xiong}]{ctrlsum}
Junxian He, Wojciech Kryscinski, Bryan McCann, Nazneen Rajani, and Caiming Xiong. 2022.
\newblock \href {https://aclanthology.org/2022.emnlp-main.396} {{CTRL}sum: Towards generic controllable text summarization}.
\newblock In \emph{Proceedings of the 2022 Conference on Empirical Methods in Natural Language Processing}, pages 5879--5915, Abu Dhabi, United Arab Emirates. Association for Computational Linguistics.

\bibitem[{He et~al.(2021)He, Gao, and Chen}]{deberta-v3}
Pengcheng He, Jianfeng Gao, and Weizhu Chen. 2021.
\newblock \href {http://arxiv.org/abs/2111.09543} {Debertav3: Improving deberta using electra-style pre-training with gradient-disentangled embedding sharing}.

\bibitem[{Hu and Liu(2004)}]{aspect-based-summarization-source-1}
Minqing Hu and Bing Liu. 2004.
\newblock \href {https://doi.org/10.1145/1014052.1014073} {Mining and summarizing customer reviews}.
\newblock In \emph{Proceedings of the Tenth ACM SIGKDD International Conference on Knowledge Discovery and Data Mining}, KDD '04, page 168–177, New York, NY, USA. Association for Computing Machinery.

\bibitem[{Keung et~al.(2020)Keung, Lu, Szarvas, and Smith}]{amazon-dataset}
Phillip Keung, Yichao Lu, Gy{\"o}rgy Szarvas, and Noah~A. Smith. 2020.
\newblock \href {https://doi.org/10.18653/v1/2020.emnlp-main.369} {The multilingual {A}mazon reviews corpus}.
\newblock In \emph{Proceedings of the 2020 Conference on Empirical Methods in Natural Language Processing (EMNLP)}, pages 4563--4568, Online. Association for Computational Linguistics.

\bibitem[{Kulkarni et~al.(2021)Kulkarni, Mahata, Arora, and Bhowmik}]{keyphrase-1}
Mayank Kulkarni, Debanjan Mahata, Ravneet Arora, and Rajarshi Bhowmik. 2021.
\newblock Learning rich representation of keyphrases from text.
\newblock \emph{arXiv preprint arXiv:2112.08547}.

\bibitem[{Kwon et~al.(2023)Kwon, Xie, Bullard, and Sadigh}]{gpt3-as-reward}
Minae Kwon, Sang~Michael Xie, Kalesha Bullard, and Dorsa Sadigh. 2023.
\newblock Reward design with language models.
\newblock \emph{arXiv preprint arXiv:2303.00001}.

\bibitem[{Laurer et~al.(2023)Laurer, Van~Atteveldt, Casas, and Welbers}]{zeroshot-nli}
Moritz Laurer, Wouter Van~Atteveldt, Andreu Casas, and Kasper Welbers. 2023.
\newblock \href {https://doi.org/10.1017/pan.2023.20} {Less {Annotating}, {More} {Classifying}: {Addressing} the {Data} {Scarcity} {Issue} of {Supervised} {Machine} {Learning} with {Deep} {Transfer} {Learning} and {BERT}-{NLI}}.
\newblock \emph{Political Analysis}, pages 1--33.

\bibitem[{Lewis et~al.(2019)Lewis, Liu, Goyal, Ghazvininejad, Mohamed, Levy, Stoyanov, and Zettlemoyer}]{bart}
Mike Lewis, Yinhan Liu, Naman Goyal, Marjan Ghazvininejad, Abdelrahman Mohamed, Omer Levy, Ves Stoyanov, and Luke Zettlemoyer. 2019.
\newblock \href {http://arxiv.org/abs/1910.13461} {Bart: Denoising sequence-to-sequence pre-training for natural language generation, translation, and comprehension}.

\bibitem[{Lin(2004)}]{rouge}
Chin-Yew Lin. 2004.
\newblock Rouge: A package for automatic evaluation of summaries.
\newblock In \emph{Text summarization branches out}, pages 74--81.

\bibitem[{Liu et~al.(2022)Liu, Swayamdipta, Smith, and Choi}]{wanli}
Alisa Liu, Swabha Swayamdipta, Noah~A. Smith, and Yejin Choi. 2022.
\newblock \href {https://doi.org/10.18653/v1/2022.findings-emnlp.508} {{WANLI}: Worker and {AI} collaboration for natural language inference dataset creation}.
\newblock In \emph{Findings of the Association for Computational Linguistics: EMNLP 2022}, pages 6826--6847, Abu Dhabi, United Arab Emirates. Association for Computational Linguistics.

\bibitem[{Nallapati et~al.(2016)Nallapati, Zhou, dos Santos, Gul{\c{c}}ehre, and Xiang}]{cnn-daily-mail}
Ramesh Nallapati, Bowen Zhou, Cicero dos Santos, {\c{C}}a{\u{g}}lar Gul{\c{c}}ehre, and Bing Xiang. 2016.
\newblock \href {https://doi.org/10.18653/v1/K16-1028} {Abstractive text summarization using sequence-to-sequence {RNN}s and beyond}.
\newblock In \emph{Proceedings of the 20th {SIGNLL} Conference on Computational Natural Language Learning}, pages 280--290, Berlin, Germany. Association for Computational Linguistics.

\bibitem[{Narayan et~al.(2018)Narayan, Cohen, and Lapata}]{xsum}
Shashi Narayan, Shay~B. Cohen, and Mirella Lapata. 2018.
\newblock \href {https://doi.org/10.18653/v1/D18-1206} {Don{'}t give me the details, just the summary! topic-aware convolutional neural networks for extreme summarization}.
\newblock In \emph{Proceedings of the 2018 Conference on Empirical Methods in Natural Language Processing}, pages 1797--1807, Brussels, Belgium. Association for Computational Linguistics.

\bibitem[{Ouyang et~al.(2022)Ouyang, Wu, Jiang, Almeida, Wainwright, Mishkin, Zhang, Agarwal, Slama, Ray, Schulman, Hilton, Kelton, Miller, Simens, Askell, Welinder, Christiano, Leike, and Lowe}]{instruct-gpt}
Long Ouyang, Jeffrey Wu, Xu~Jiang, Diogo Almeida, Carroll Wainwright, Pamela Mishkin, Chong Zhang, Sandhini Agarwal, Katarina Slama, Alex Ray, John Schulman, Jacob Hilton, Fraser Kelton, Luke Miller, Maddie Simens, Amanda Askell, Peter Welinder, Paul~F Christiano, Jan Leike, and Ryan Lowe. 2022.
\newblock \href {https://proceedings.neurips.cc/paper_files/paper/2022/file/b1efde53be364a73914f58805a001731-Paper-Conference.pdf} {Training language models to follow instructions with human feedback}.
\newblock In \emph{Advances in Neural Information Processing Systems}, volume~35, pages 27730--27744. Curran Associates, Inc.

\bibitem[{Pasunuru and Bansal(2018)}]{rouge-rl-summarization-1}
Ramakanth Pasunuru and Mohit Bansal. 2018.
\newblock \href {https://doi.org/10.18653/v1/N18-2102} {Multi-reward reinforced summarization with saliency and entailment}.
\newblock In \emph{Proceedings of the 2018 Conference of the North {A}merican Chapter of the Association for Computational Linguistics: Human Language Technologies, Volume 2 (Short Papers)}, pages 646--653, New Orleans, Louisiana. Association for Computational Linguistics.

\bibitem[{Paszke et~al.(2019)Paszke, Gross, Massa, Lerer, Bradbury, Chanan, Killeen, Lin, Gimelshein, Antiga, Desmaison, Kopf, Yang, DeVito, Raison, Tejani, Chilamkurthy, Steiner, Fang, Bai, and Chintala}]{PyTorch}
Adam Paszke, Sam Gross, Francisco Massa, Adam Lerer, James Bradbury, Gregory Chanan, Trevor Killeen, Zeming Lin, Natalia Gimelshein, Luca Antiga, Alban Desmaison, Andreas Kopf, Edward Yang, Zachary DeVito, Martin Raison, Alykhan Tejani, Sasank Chilamkurthy, Benoit Steiner, Lu~Fang, Junjie Bai, and Soumith Chintala. 2019.
\newblock \href {http://papers.neurips.cc/paper/9015-pytorch-an-imperative-style-high-performance-deep-learning-library.pdf} {Pytorch: An imperative style, high-performance deep learning library}.
\newblock In \emph{Advances in Neural Information Processing Systems 32}, pages 8024--8035. Curran Associates, Inc.

\bibitem[{Paulus et~al.(2018)Paulus, Xiong, and Socher}]{rouge-rl-summarization-2}
Romain Paulus, Caiming Xiong, and Richard Socher. 2018.
\newblock \href {https://openreview.net/forum?id=HkAClQgA-} {A deep reinforced model for abstractive summarization}.
\newblock In \emph{6th International Conference on Learning Representations, {ICLR} 2018, Vancouver, BC, Canada, April 30 - May 3, 2018, Conference Track Proceedings}. OpenReview.net.

\bibitem[{Radford et~al.(2019)Radford, Wu, Child, Luan, Amodei, Sutskever et~al.}]{gpt-2}
Alec Radford, Jeffrey Wu, Rewon Child, David Luan, Dario Amodei, Ilya Sutskever, et~al. 2019.
\newblock Language models are unsupervised multitask learners.
\newblock \emph{OpenAI blog}, 1(8):9.

\bibitem[{Sahrawat et~al.(2020)Sahrawat, Mahata, Zhang, Kulkarni, Sharma, Gosangi, Stent, Kumar, Shah, and Zimmermann}]{keyphrase-2}
Dhruva Sahrawat, Debanjan Mahata, Haimin Zhang, Mayank Kulkarni, Agniv Sharma, Rakesh Gosangi, Amanda Stent, Yaman Kumar, Rajiv~Ratn Shah, and Roger Zimmermann. 2020.
\newblock Keyphrase extraction as sequence labeling using contextualized embeddings.
\newblock In \emph{Advances in Information Retrieval: 42nd European Conference on IR Research, ECIR 2020, Lisbon, Portugal, April 14--17, 2020, Proceedings, Part II 42}, pages 328--335. Springer.

\bibitem[{Schluter(2017)}]{rouge-limits}
Natalie Schluter. 2017.
\newblock \href {https://aclanthology.org/E17-2007} {The limits of automatic summarisation according to {ROUGE}}.
\newblock In \emph{Proceedings of the 15th Conference of the {E}uropean Chapter of the Association for Computational Linguistics: Volume 2, Short Papers}, pages 41--45, Valencia, Spain. Association for Computational Linguistics.

\bibitem[{Schulman et~al.(2017)Schulman, Wolski, Dhariwal, Radford, and Klimov}]{ppo}
John Schulman, Filip Wolski, Prafulla Dhariwal, Alec Radford, and Oleg Klimov. 2017.
\newblock Proximal policy optimization algorithms.
\newblock \emph{arXiv preprint arXiv:1707.06347}.

\bibitem[{See et~al.(2017)See, Liu, and Manning}]{Pointer-Generator}
Abigail See, Peter~J. Liu, and Christopher~D. Manning. 2017.
\newblock \href {https://doi.org/10.18653/v1/P17-1099} {Get to the point: Summarization with pointer-generator networks}.
\newblock In \emph{Proceedings of the 55th Annual Meeting of the Association for Computational Linguistics (Volume 1: Long Papers)}, pages 1073--1083, Vancouver, Canada. Association for Computational Linguistics.

\bibitem[{Song et~al.(2020)Song, Shuai, Yeh, Wu, Ku, and Peng}]{Attractive-AAAI2020}
Yun-Zhu Song, Hong-Han Shuai, Sung-Lin Yeh, Yi-Lun Wu, Lun-Wei Ku, and Wen-Chih Peng. 2020.
\newblock \href {https://doi.org/10.1609/aaai.v34i05.6421} {Attractive or faithful? popularity-reinforced learning for inspired headline generation}.
\newblock \emph{Proceedings of the AAAI Conference on Artificial Intelligence}, 34(05):8910--8917.

\bibitem[{Stiennon et~al.(2020)Stiennon, Ouyang, Wu, Ziegler, Lowe, Voss, Radford, Amodei, and Christiano}]{RLHF-summarization}
Nisan Stiennon, Long Ouyang, Jeffrey Wu, Daniel Ziegler, Ryan Lowe, Chelsea Voss, Alec Radford, Dario Amodei, and Paul~F Christiano. 2020.
\newblock \href {https://proceedings.neurips.cc/paper_files/paper/2020/file/1f89885d556929e98d3ef9b86448f951-Paper.pdf} {Learning to summarize with human feedback}.
\newblock In \emph{Advances in Neural Information Processing Systems}, volume~33, pages 3008--3021. Curran Associates, Inc.

\bibitem[{Sun et~al.(2017)Sun, Ajwani, Nicholson, Sala, and Parthasarathy}]{break-cycle}
Jiankai Sun, Deepak Ajwani, Patrick~K. Nicholson, Alessandra Sala, and Srinivasan Parthasarathy. 2017.
\newblock \href {https://doi.org/10.1145/3091478.3091495} {Breaking cycles in noisy hierarchies}.
\newblock In \emph{Proceedings of the 2017 ACM on Web Science Conference}, WebSci '17, pages 151--160, New York, NY, USA. ACM.

\bibitem[{Tas and Kiyani(2007)}]{summarization-survey-1}
Oguzhan Tas and Farzad Kiyani. 2007.
\newblock A survey automatic text summarization.
\newblock \emph{PressAcademia Procedia}, 5(1):205--213.

\bibitem[{Touvron et~al.(2023)Touvron, Martin, Stone, Albert, Almahairi, Babaei, Bashlykov, Batra, Bhargava, Bhosale et~al.}]{llama2}
Hugo Touvron, Louis Martin, Kevin Stone, Peter Albert, Amjad Almahairi, Yasmine Babaei, Nikolay Bashlykov, Soumya Batra, Prajjwal Bhargava, Shruti Bhosale, et~al. 2023.
\newblock Llama 2: Open foundation and fine-tuned chat models.
\newblock \emph{arXiv preprint arXiv:2307.09288}.

\bibitem[{von Werra et~al.(2020)von Werra, Belkada, Tunstall, Beeching, Thrush, and Lambert}]{trl}
Leandro von Werra, Younes Belkada, Lewis Tunstall, Edward Beeching, Tristan Thrush, and Nathan Lambert. 2020.
\newblock Trl: Transformer reinforcement learning.
\newblock \url{https://github.com/lvwerra/trl}.

\bibitem[{Wang et~al.(2018)Wang, Singh, Michael, Hill, Levy, and Bowman}]{glue}
Alex Wang, Amanpreet Singh, Julian Michael, Felix Hill, Omer Levy, and Samuel~R Bowman. 2018.
\newblock Glue: A multi-task benchmark and analysis platform for natural language understanding.
\newblock \emph{arXiv preprint arXiv:1804.07461}.

\bibitem[{Wang et~al.(2017)Wang, Hou, Liu, Cao, and Lin}]{product-description-generation}
Jinpeng Wang, Yutai Hou, Jing Liu, Yunbo Cao, and Chin-Yew Lin. 2017.
\newblock A statistical framework for product description generation.
\newblock In \emph{Proceedings of the Eighth International Joint Conference on Natural Language Processing (Volume 2: Short Papers)}, pages 187--192.

\bibitem[{Wang and Wan(2018)}]{perplexity-fluency-1}
Ke~Wang and Xiaojun Wan. 2018.
\newblock Sentigan: Generating sentimental texts via mixture adversarial networks.
\newblock In \emph{IJCAI}, pages 4446--4452.

\bibitem[{Williams et~al.(2018)Williams, Nangia, and Bowman}]{mnli}
Adina Williams, Nikita Nangia, and Samuel Bowman. 2018.
\newblock \href {https://doi.org/10.18653/v1/N18-1101} {A broad-coverage challenge corpus for sentence understanding through inference}.
\newblock In \emph{Proceedings of the 2018 Conference of the North {A}merican Chapter of the Association for Computational Linguistics: Human Language Technologies, Volume 1 (Long Papers)}, pages 1112--1122, New Orleans, Louisiana. Association for Computational Linguistics.

\bibitem[{Wolf et~al.(2019)Wolf, Debut, Sanh, Chaumond, Delangue, Moi, Cistac, Rault, Louf, Funtowicz, and Brew}]{Huggingface}
Thomas Wolf, Lysandre Debut, Victor Sanh, Julien Chaumond, Clement Delangue, Anthony Moi, Pierric Cistac, Tim Rault, R{\'{e}}mi Louf, Morgan Funtowicz, and Jamie Brew. 2019.
\newblock \href {http://arxiv.org/abs/1910.03771} {Huggingface's transformers: State-of-the-art natural language processing}.
\newblock \emph{CoRR}, abs/1910.03771.

\bibitem[{Wu et~al.(2016)Wu, Gu, Sun, and Gu}]{Aspect-Opinion-Summarization-1}
Haibing Wu, Yiwei Gu, Shangdi Sun, and Xiaodong Gu. 2016.
\newblock \href {https://doi.org/10.1109/IJCNN.2016.7727602} {Aspect-based opinion summarization with convolutional neural networks}.
\newblock In \emph{2016 International Joint Conference on Neural Networks (IJCNN)}, pages 3157--3163.

\bibitem[{Xiao and Munro(2019)}]{product-titles}
Joan Xiao and Robert Munro. 2019.
\newblock Text summarization of product titles.
\newblock In \emph{eCOM@ SIGIR}.

\bibitem[{Xu et~al.(2019)Xu, Wu, Madotto, and Fung}]{clickbait}
Peng Xu, Chien-Sheng Wu, Andrea Madotto, and Pascale Fung. 2019.
\newblock \href {https://doi.org/10.18653/v1/D19-1303} {Clickbait? sensational headline generation with auto-tuned reinforcement learning}.
\newblock In \emph{Proceedings of the 2019 Conference on Empirical Methods in Natural Language Processing and the 9th International Joint Conference on Natural Language Processing (EMNLP-IJCNLP)}, pages 3065--3075, Hong Kong, China. Association for Computational Linguistics.

\bibitem[{Yang et~al.(2023)Yang, Li, Zhang, Chen, and Cheng}]{chatgpt-aspect-summarization-limitation}
Xianjun Yang, Yan Li, Xinlu Zhang, Haifeng Chen, and Wei Cheng. 2023.
\newblock \href {http://arxiv.org/abs/2302.08081} {Exploring the limits of chatgpt for query or aspect-based text summarization}.

\bibitem[{Zhang et~al.(2018)Zhang, Guo, Fan, Lan, Xu, Cao, and Cheng}]{question-headline-generation}
Ruqing Zhang, Jiafeng Guo, Yixing Fan, Yanyan Lan, Jun Xu, Huanhuan Cao, and Xueqi Cheng. 2018.
\newblock \href {https://doi.org/10.1145/3269206.3271711} {Question headline generation for news articles}.
\newblock In \emph{Proceedings of the 27th ACM International Conference on Information and Knowledge Management}, CIKM '18, page 617–626, New York, NY, USA. Association for Computing Machinery.

\bibitem[{Zhang et~al.(2023)Zhang, Ladhak, Durmus, Liang, McKeown, and Hashimoto}]{chatgpt-llm-benchmark}
Tianyi Zhang, Faisal Ladhak, Esin Durmus, Percy Liang, Kathleen McKeown, and Tatsunori~B. Hashimoto. 2023.
\newblock \href {http://arxiv.org/abs/2301.13848} {Benchmarking large language models for news summarization}.

\bibitem[{Zhang et~al.(2022)Zhang, Zou, Zhang, Zhou, Diao, Chen, Ding, He, He, Xiao et~al.}]{copywriting}
Xueying Zhang, Yanyan Zou, Hainan Zhang, Jing Zhou, Shiliang Diao, Jiajia Chen, Zhuoye Ding, Zhen He, Xueqi He, Yun Xiao, et~al. 2022.
\newblock Automatic product copywriting for e-commerce.
\newblock In \emph{Proceedings of the AAAI Conference on Artificial Intelligence}, volume~36, pages 12423--12431.

\bibitem[{Zhao et~al.(2020)Zhao, Walker, and Chaturvedi}]{Human-evaluation-setting}
Chao Zhao, Marilyn Walker, and Snigdha Chaturvedi. 2020.
\newblock \href {https://doi.org/10.18653/v1/2020.acl-main.224} {Bridging the structural gap between encoding and decoding for data-to-text generation}.
\newblock In \emph{Proceedings of the 58th Annual Meeting of the Association for Computational Linguistics}, pages 2481--2491, Online. Association for Computational Linguistics.

\bibitem[{Ziegler et~al.(2019)Ziegler, Stiennon, Wu, Brown, Radford, Amodei, Christiano, and Irving}]{RLHF-2019-openai}
Daniel~M Ziegler, Nisan Stiennon, Jeffrey Wu, Tom~B Brown, Alec Radford, Dario Amodei, Paul Christiano, and Geoffrey Irving. 2019.
\newblock Fine-tuning language models from human preferences.
\newblock \emph{arXiv preprint arXiv:1909.08593}.

\bibitem[{Zulkifly and Firdaus(2014)}]{zulkifly2014persuasion}
Hani~Zulaikha Zulkifly and Norsham Firdaus. 2014.
\newblock Persuasion and the online consumers: Investigating copywriting strategies in native advertisements.
\newblock \emph{International Journal of Social Science and Humanity}, 4(6):430.

\end{thebibliography}

\appendix

\section{Implementation Details}
\label{sec:implementation_details}
\paragraph{Allure Reward Model}
For the training of the allure RM, we adopt a debate-v3 model as the backbone and set the batch size to 32, with a learning rate of 2e-5. Since the regression model and Siamese network converge earlier, we train them for 5 epochs.

\paragraph{Information Reward Model}
Since the QNLI dataset is included in a well-known benchmark named GLUE \citep{glue}, there are numerous off-the-shelf models provided in the huggingface platform \citep{Huggingface}; we select one of them for all experiments\footnote{https://huggingface.co/cross-encoder/qnli-electra-base}. The reported accuracy of this model on the QNLI dev set is 93.21\%. 

\paragraph{Reinforcement Learning}
We adopt the trl implementation provided by the huggingface platform \citep{trl} for this part. The learning rate is set to be 1e-6, and the batch size is set to be 32. We fine-tune the model for 20 epochs. Although it seems that there is still room for enhancement. We observe instability in the training process after 20 epochs and might be unable to reproduce sometimes. The training procedure takes around 12 hours on a single Nvidia Tesla-V100 GPU.

\section{Queries for Information Reward}
\label{sec:queries}
Given an aspect $k$,  we define some common facets of restaurant review as following:
\renewcommand\labelitemi{\ding{117}}

\begin{itemize}
    \item What is the price of $k$?
    \vspace{-1ex}
    \item What is the cooking method of $k$?
    \vspace{-1ex}
    \item How does the $k$ taste?
    \vspace{-1ex}
    \item How does the $k$ smell?
    \vspace{-1ex}
    \item How does the $k$ look?
    \vspace{-1ex}
    \item How is the quality of $k$?
    \vspace{-1ex}
    \item How is the seasoning of $k$?
    \vspace{-1ex}
    \item How is the portion size of $k$ ?
	\vspace{-1ex}
    \item How is the cleanliness of $k$
    \vspace{-1ex}	
    \item What are the components of the $k$?
    \vspace{-1ex}	
    \item What is the opinion of the people about the $k$?
    \vspace{-4ex}	
    \item How does $k$ compare to similar ones on the market?

\end{itemize}

Despite the template-based queries not being suitable for all scenarios, it's reasonable to let the model handle itself because there is no "answer" to unreasonable queries in the world. According to our preliminary experiments, the QNLI model indeed owns this ability.

\section{Generated Examples}
\label{app:generated_example_full}

The examples generated by each baseline model are shown in Table \ref{tab:generated_example_full}.
\label{sec:generated_example}
\begin{table*}[h]
\centering
\scalebox{1}{
\begin{tabular}{l|>{\arraybackslash}p{6cm}|>{\arraybackslash}p{6cm}}
\toprule[1pt] 
    \midrule
     \multicolumn{1}{c|}{Aspect} & \multicolumn{1}{c|}{Steak} & \multicolumn{1}{c}{Tampa} \\
\midrule
     Source Review      & \multicolumn{2}{>{\arraybackslash}p{12.5cm}}{We got diverted to Tampa and decided to make the drive to Orlando, but first we needed somewhere to eat. Wanting to avoid chain restaurants,  yelp steered us  toward this little local gem! They grill their steaks over oak, and citrus Wood  and the smoky flavor is to die for! We are big fans of our smoker back home  and cook our steaks this way as well. It's rare to find steak cooked over woodsmoke  in most places, and this was worth the side trip!  The sides were veggies and  sweet potatoes. The wines on offer were so reasonably priced - most under \$20 a  bottle, but a decent selection. They just don't mark them up 5x like most places!  Bookmarking this place for a return trip next time I am in Tampa for sure!}   \\ 

    \midrule[1pt]
    \midrule

     GPT-3.5      & Juicy, tender, and perfectly grilled, this oak and citrus wood-smoked steak delivers a mouthwatering  smoky flavor. Cooked to perfection, it's a rare  gem worth savoring, a true delight for steak enthusiasts. &  An enchanting city in Florida known for its diverse dining options, Tampa captivates visitors with its local gems and the tantalizing aroma of woodsmoked steaks
     \\
    
    \midrule
    
     LLaMA-2-7B-Chat & Savory, smoky steak grilled over oak and citrus wood, a true delight for steak lovers.    &  Tampa's hidden gem for steak lovers:  wood-grilled masterpieces in a cozy local setting.  \\ 
    \midrule 
    
    
     SFT               & Smoky, charred steak cooked over oak and citrus wood, a rare delicacy in Tampa. Worth the side trip!   & Tampa: A local gem with smoky steaks over  oak and citrus wood , delicious sides, affordable wines, and exceptional service. A must-visit!  \\    

    \midrule
    
     CTRLsum           & Mouthwatering, tender, and smoky steaks grilled over oak and citrus Wood,  bursting with a delectable flavor  that will leave you craving for more. &  Vibrant city with diverse culinary options,  including a local gem serving smoky steaks  and delicious sides.
     \\

    \midrule
    
     ROUGE             & Juicy, smoky, and grilled over oak and citrus wood, our   steaks at this local gem in Tampa are truly exceptional.   A rare and mouthwatering delight cooked over smoky  flavor,  worth the side trip and the friendly service. & Vibrant Florida city with smoky steaks, affordable wines, and delicious sides.  A must-visit destination.  \\

    \midrule
    
     Ours              & Juicy, smoky, and grilled over oak and citrus wood, our steaks at this local gem in Tampa are truly exceptional.  A rare delight cooked over woodsmoke,  it's worth the  side trip! The sides were veggies and sweet potatoes.  &  Vibrant Florida city with smoky steaks, affordable wines,  delicious veggies  and sweet potatoes.  A must-visit!    \\
\midrule 
\bottomrule[1pt]
\end{tabular}
}
\caption{The generated examples of different models.}
\label{tab:generated_example_full}
\end{table*}

\end{document}